\pdfoutput=1

\documentclass[11pt]{article}

\usepackage[final]{acl}

\usepackage{times}
\usepackage{latexsym}
\usepackage{amsmath} 
\usepackage{float} 
\usepackage{enumitem}
\setlist{nosep} 

\usepackage[T1]{fontenc}

\usepackage[utf8]{inputenc}

\usepackage{microtype}
\usepackage{multirow}
\usepackage{subcaption}

\usepackage{inconsolata}

\usepackage{graphicx}

\title{A Two-Model Approach for Humour Style Recognition}

\author{
  \textbf{Mary Ogbuka Kenneth\textsuperscript{1}},
  \textbf{Foaad Khosmood\textsuperscript{2}},
  \textbf{Abbas Edalat\textsuperscript{1}} \\
  \textsuperscript{1}Algorithmic Human Development group, Department of Computing, Imperial College London, UK \\
  \textsuperscript{2}Computer Engineering Department, California Polytechnic State University, USA \\
  \texttt{m.kenneth22@imperial.ac.uk, foaad@calpoly.edu, a.edalat@imperial.ac.uk}
}

\begin{document}
\maketitle
\begin{abstract}
Humour, a fundamental aspect of human communication, manifests itself in various styles that significantly impact social interactions and mental health. Recognising different humour styles poses challenges due to the lack of established datasets and machine learning (ML) models. To address this gap, we present a new text dataset for humour style recognition, comprising 1463 instances across four styles (self-enhancing, self-deprecating, affiliative, and aggressive) and non-humorous text, with lengths ranging from 4 to 229 words. Our research employs various computational methods, including classic machine learning classifiers, text embedding models, and DistilBERT, to establish baseline performance. Additionally, we propose a two-model approach to enhance humour style recognition, particularly in distinguishing between affiliative and aggressive styles. Our method demonstrates an 11.61\% improvement in f1-score for affiliative humour classification, with consistent improvements in the 14 models tested. Our findings contribute to the computational analysis of humour in text, offering new tools for studying humour in literature, social media, and other textual sources.
\end{abstract}

\section{Introduction}
Humour recognition is a multidimensional task influenced by various theories and manifested through diverse styles. There are various humour theories, such as relief, incongruity, and superiority theories \citep{Morreall2011ComicHumor,Morreall2012PhilosophyHumor,Scheel2017HumorHealth}. The relief theory highlights the role of humour in relaxation, while the incongruity theory suggests that we find something funny when we notice a mismatch or contradiction between what we expect in a situation and what actually happens. The superiority theory suggests that people may laugh at other people's misfortunes in an effort to demonstrate their superiority.

These theories not only explain why we find things humorous but also why we laugh as a response. In recent decades, evolutionary psychology has introduced a new perspective on laughter itself, known as the {\em play} theory \citep{Martin2018TheApproach}: laughter developed as a play signal in higher primates in their mock fights to indicate non-aggressive intent.

Laughter, therefore, is more than just a reaction to humour; it serves various functions, including promoting mental, emotional, and physical well-being. This idea forms the basis for laughter therapy, a cognitive-behavioural treatment designed to induce laughter and reduce stress, tension, anxiety, and sadness \citep{Yim2016TherapeuticReview}. However, as \citet{Martin2003IndividualQuestionnaire} noted, not all humour is beneficial—some forms can even harm relationships with others or oneself.

Considering its impact on well-being, \citet{Martin2003IndividualQuestionnaire} categorised humour into four styles: self-enhancing, self-deprecating, affiliative, and aggressive. Affiliative and self-enhancing humour are beneficial to psychological well-being. Affiliative humour fosters social bonding, while self-enhancing humour involves maintaining a positive outlook without harming oneself or others, often employed as a coping mechanism in difficult situations \citep{Edalat2023Self-initiatedLaugh,Kenneth2024SystematicClassification,Hampes2007TheEmpathy,Plessen2020HumorTraits}. In contrast, aggressive and self-deprecating humour can be harmful. Aggressive humour, rooted in superiority theory, belittles or mocks others, whereas self-deprecating humour seeks approval by making oneself the target of jokes \citep{Khramtsova2016MindfulnessRussia,Kuiper2016IdentityWell-Being,Veselka2010RelationsPersonality}.

In artificial intelligence (AI), humour is considered AI-complete \citep{Shani2021HowAchievements,Strapparava2011ComputationalHumour,Kenneth2024SystematicClassification}, meaning that a system capable of producing and recognising human-like humour would possess general intelligence. Despite the importance of humour, most computational efforts have focused on laughter detection \citep{Vargas-Quiros2023ImpactIn-the-Wild,Matsuda2023DetectionModels,Inoue2022CanDialogue}, classification \citep{Tanaka2014ClassificationSpeech} and generation \citep{Inoue2022CanDialogue}, as well as humour detection \citep{Oliveira2020CorporaPortuguese,Jaiswal2019AutomaticParadigms,ChauhanAll-in-One:Memes}, and humour generation \citep{Luo2019Pun-GAN:Generation,He2019PunSurprise,Yu2018AGeneration}, with little emphasis on humour styles and their links to well-being. \citet{Kenneth2024SystematicClassification} identified a gap in the current ML landscape: the lack of datasets and models specifically designed to recognise these four humour styles.

Building on the gaps identified by \citet{Kenneth2024SystematicClassification}, this study addresses the lack of an established dataset and ML models for recognising the four humour styles: self-enhancing, self-deprecating, affiliative, and aggressive. We draw on \citet{Martin2003IndividualQuestionnaire}, who defined and validated these styles, providing the theoretical basis for our classification task. Additionally, \citet{Edalat2023Self-initiatedLaugh}'s work on self-initiated humour protocols (SIHP) informs how different humour styles can enhance well-being, while \citet{Amjad2022HumorAdults} research on the link between humour styles, emotion regulation, and subjective well-being highlights the potential applications of our work in psychological and clinical contexts. By integrating these insights, we aim to develop a comprehensive approach to humour style recognition grounded in psychological theory and applicable to real-world scenarios. The key contributions of this paper are:

\begin{enumerate}
    \item  Introduction of a new text dataset for humour style recognition, addressing the lack of established datasets. This dataset is publicly available to the community. 
    \item  Baseline evaluations using various ML classifiers and models.
    \item  Development of a two-model approach for improved humour style recognition.
    \item  Extensive evaluation of the proposed two-model approach. 
\end{enumerate}

\section{Related Works}
Humour recognition and classification are active research areas in NLP and multi-modal analysis. While our focus is on humour style recognition, we draw insights from related fields like general humour detection and sarcasm detection.

\citet{Weller2020TheCollection} compiled a dataset of 550,000 jokes from Reddit posts, using user ratings and engagement metrics as quantifiable humour quality measurements. However, the dataset’s reliance on Reddit data alone may introduce biases and limit generalisability. Our study addresses this by introducing a more diverse dataset specifically tailored for humour style recognition from various online sources.

\citet{Oliveira2020CorporaPortuguese} explored humour recognition in Portuguese text, achieving a 75\% f1-score using Naive Bayes, Support Vector Machine, and Random Forest classifiers. However, their work was limited to binary classification of headlines and one-liners. Our approach extends this by focusing on multi-class classification of humour styles in both short and long texts.

\citet{Tang2022TheHumor} created a dataset and classification model for sub-types of inappropriate humour, using large language models like BERT. While relevant, their focus on inappropriate humour differs from our goal of recognising humour styles linked to psychological well-being.

\citet{Kamal2020Self-deprecatingApproach} targeted self-deprecating humour, one of the four styles we examine. Their use of specific feature categories (self-deprecating pattern, and word embedding) informs our feature engineering process. However, our study broadens the scope to include all four humour styles.

\citet{Christ2022TheStress,Christ2022MultimodalResults} developed models for humour recognition in German football press conferences. Although their work yielded promising results, it was limited to the MuSe humour challenge and the Passau-SFCH German dataset, unlike our broader approach.

Sarcasm detection is closely related to humour style recognition since it is often used in aggressive and self-deprecating humour styles. \citet{Liang2021Multi-ModalGraphs} used an interactive graph convolution network for multi-modal sarcasm detection, highlighting the importance of contextual cues. This technique could be adapted to distinguish humour styles.

\citet{Jinks2023IntermediateDetection} improved sarcasm detection with a two-step fine-tuning process using RoBERTa, a method that could enhance humour style classification given the subtle differences between styles.

\citet{Fang2024Single-StageDetection} introduces the Single-Stage Extensive Semantic Fusion model for multi-modal sarcasm detection by concurrently processing and fusing multi-modal inputs in a unified framework. This approach could be adapted for humour style recognition, when we expand our dataset to include multi-modal features in the future.

Although these studies contribute to the detection of humour and sarcasm, there is a gap in recognising the four humour styles defined by \citet{Martin2003IndividualQuestionnaire}. Our work fills this gap by creating a dedicated dataset and developing classification models tailored to these humour styles.

\section{Dataset Collection and Annotation}
A significant challenge in identifying humour styles automatically is the lack of annotated datasets suitable for training machine learning models. To address this, we created a comprehensive dataset comprising 1,463 instances from various sources:
\begin{enumerate}
    \item 983 jokes from several well-known websites where jokes were labelled by users or editors.
    \item 280 non-humorous text instances from the ColBERT dataset \citep{Annamoradnejad2020ColBERT:Humor}.
    \item 200 instances from the Short Text Corpus \footnote{Short Text Corpus (\url{https://github.com/CrowdTruth/Short-Text-Corpus-For-Humor-Detection})}, consisting of 150 jokes and 50 non-jokes
\end{enumerate}

After annotation, the dataset consists of 298 instances of self-enhancing humour, 265 of self-deprecating humour, 250 of affiliative humour, 318 of aggressive humour, and 332 neutral instances, with text lengths ranging from 4 to 229 words. This distribution ensures balanced representation across the different humour styles and neutral text.

\subsection{Data Sources and Labelling}
The 983 jokes were extracted from sources like Reader's Digest, Parade, Bored Panda, Laugh Factory, Pun Me, Independent, Cracked, Reddit, Tastefully Offensive and BuzzFeed. We labelled each joke based on the original labels, definitions, or tags given on the websites, mapping them to our categories based on humour theory. Table \ref{tab:equivalence} summarises these mappings, illustrating how the website tags correspond to our humour style labels. 

\begin{table}[h]
    \centering
    \resizebox{0.50\textwidth}{!}{
    \begin{tabular}{l|l}
    \hline
    \textbf{Equivalence Classes (Website Keywords)} & \textbf{Humour Styles} \\
    \hline
    Dark (inappropriate) Jokes & Aggressive \\
    Insult & Aggressive \\ \hline
    Icebreakers Jokes for Work Meetings & Affiliative \\
    International Day of Happiness & Affiliative \\
    Friendship & Affiliative \\
    Family jokes & Affiliative \\
    Classroom jokes & Affiliative \\ \hline
    Self-deprecating & Self-deprecating \\ \hline
    Self-love & Self-enhancing \\
    Self-care & Self-enhancing \\
    \hline
    \end{tabular}
    }
    \caption{Terminological Equivalence Classes}
    \label{tab:equivalence}
\end{table} 

For example, in Table \ref{tab:equivalence} the "Dark (inappropriate)" tag was mapped to the aggressive style because dark or inappropriate jokes are identified as being cruel, morbid, or offensive to some, which aligns with the characteristics of aggressive humour \citep{Tang2022TheHumor}. Further details on these mappings are available in Appendix \ref{appendix:mapping_labels}. 

To simulate real-life scenarios where users might input non-humorous text, we added 280 non-humorous instances from the ColBERT dataset \citep{Annamoradnejad2020ColBERT:Humor}, labelled as Neutral. 

\begin{figure}[t]
  \includegraphics[width=\columnwidth]{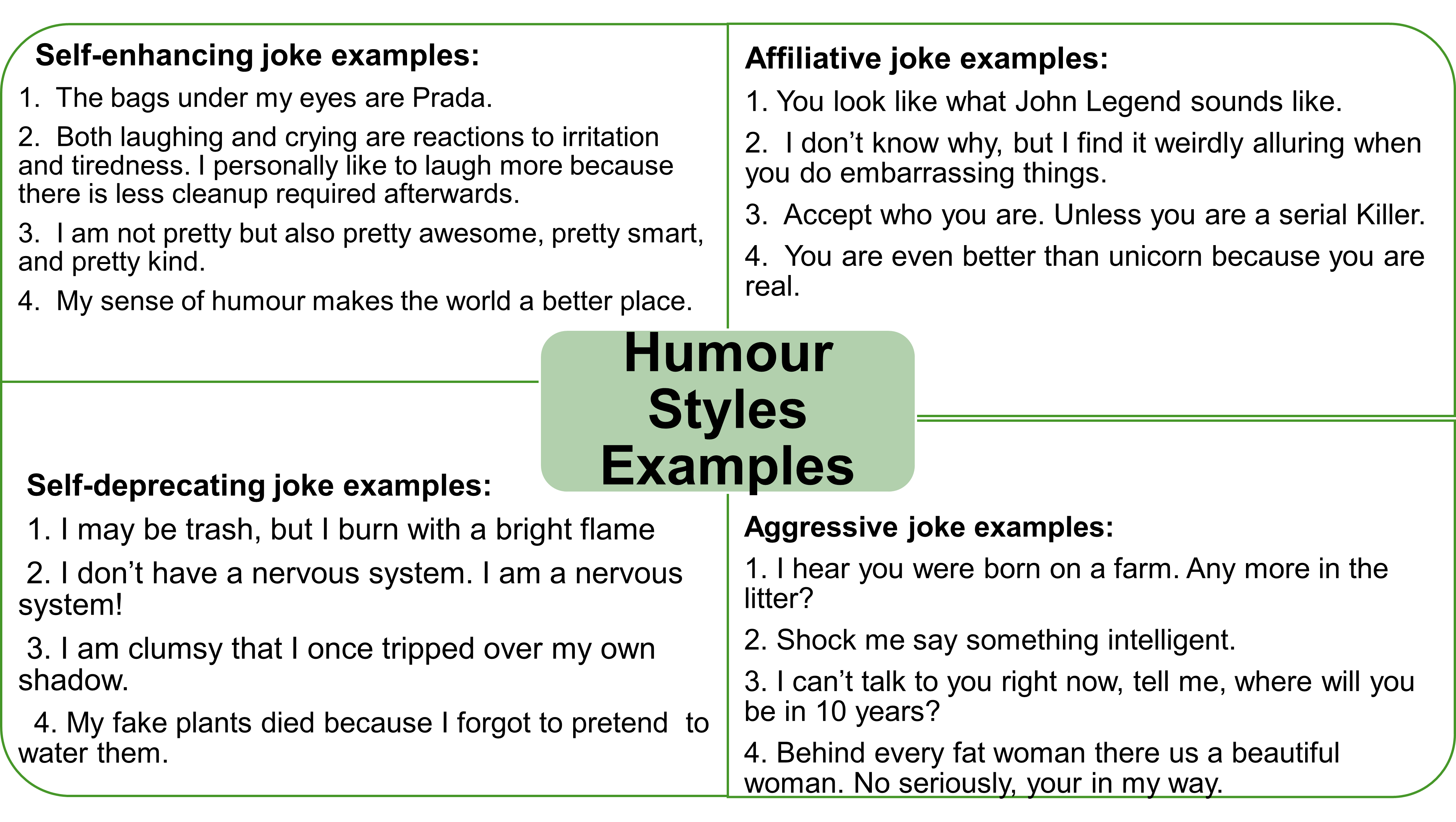}
  \caption{Joke Examples for Each Humour Style}
  \label{fig:examples_styles}
\end{figure}

Figure \ref{fig:examples_styles} presents random examples from the dataset for each humour style. Additionally, word clouds showing the most common words associated with each humour style in the created dataset are provided in Appendix \ref{appendix:word_cloud}. 

\subsection{Dataset Composition and Potential Biases}
Each humour style in our dataset was primarily sourced from different websites (see Table \ref{Table:dataset-website-links} in Appendix \ref{appendix:humour_websites} for details). The use of diverse websites, catering to various audiences and content styles, helps mitigate biases that could arise from relying on a single source. However, since the jokes were collected in English, there may be language biases, as humour often involves nuances and idioms specific to certain languages and cultures.

By aggregating data from multiple websites, we aimed to reduce inherent biases from any single source and provide comprehensive coverage of different humour styles, enhancing the robustness of the dataset. However, most websites (except Reader's Digest and Laugh Factory) featured jokes from only one humour type, potentially introducing idiosyncratic styles that could lead the classifier to learn spurious correlations.

To address this concern and further diversify our dataset, we included an additional 200 jokes from the existing Short Text Corpus joke dataset\footnotemark[1] and have them annotated by six human annotators. Details of the Short Text Corpus\footnotemark[1] and the annotation process are discussed further in the following subsection.

\subsection{Annotation Process and Inter-annotator Agreement}
Building on our efforts to address potential biases in our dataset composition, we took additional steps to ensure the robustness of our data. To mitigate potential biases from idiosyncratic styles of the individual websites, we randomly selected 200 instances from the Short Text Corpus\footnotemark[1], dividing them into two sets of 100 samples. This corpus was chosen for its diversity, featuring both short and long jokes from more than seven sources, as well as non-jokes from three sources. In contrast, the ColBERT dataset \citep{Annamoradnejad2020ColBERT:Humor} was not used here because it consists solely of Reddit jokes, which would not address the issue of spurious correlations.

To further ensure the reliability of our annotations, we recruited six Ph.D. candidates from Africa, Asia, and Europe to serve as annotators, bringing a diverse range of analytical perspectives to the task. Each set of 100 samples was independently annotated by three annotators, who were provided with humour style definitions and asked to classify each instance as self-enhancing, self-deprecating, aggressive, affiliative, or neutral. A majority vote determined the final label for each instance.

Fleiss' Kappa was used to assess inter-annotator agreement. The results showed fair agreement levels:

\begin{enumerate}
    \item First 100 samples: Fleiss' Kappa = 0.2651
    \item Second 100 samples: Fleiss' Kappa = 0.2841
\end{enumerate}

Despite the relatively low Kappa values, further analysis showed substantial agreement among at least two annotators:

\begin{enumerate}
    \item For the first set of 100 samples: 91 samples had at least two annotators agreeing on the label and 9 instances had all three annotators disagreeing.
    \item For the second set of 100 samples: 95 samples had at least two annotators agreeing on the label and 5 instances had all three annotators disagreeing.
\end{enumerate}

To resolve the 14 instances (9 in the first set, 5 in the second) where all three annotators disagreed, indicating no majority vote, we used four Large Language Models (LLMs) chatbots: \href{https://chatgpt.com/auth/login}{ChatGPT}, \href{https://claude.ai/login?returnTo=%2F%3F}{Claude}, \href{https://copilot.microsoft.com/}{Microsoft Copilot}, and \href{https://huggingface.co/chat/}{HuggingChat} - to classify the jokes. We prompted the LLMs to categorise each joke instance as self-enhancing, self-deprecating, aggressive, affiliative, or neutral. 
Each of the 14 instances then had seven labels (from the 4 LLMs and 3 human annotators), and the majority label was assigned. Table \ref{tab:annotation_disagreement} in Appendix \ref{appendix:annotation_disagreement} provides examples of instances where annotators disagreed, along with the annotators' and LLMs' labels.

These disagreements highlight the subjective nature of humour interpretation, which can be influenced by cultural differences, personal experiences, and individual preferences \citep{Lu2023CulturalCritique}. This subjectivity is a natural aspect of humour annotation, and our use of multiple annotators and LLMs helps to mitigate its impact.

\section{Methodology}
\begin{figure}[t]
  \includegraphics[width=\columnwidth]{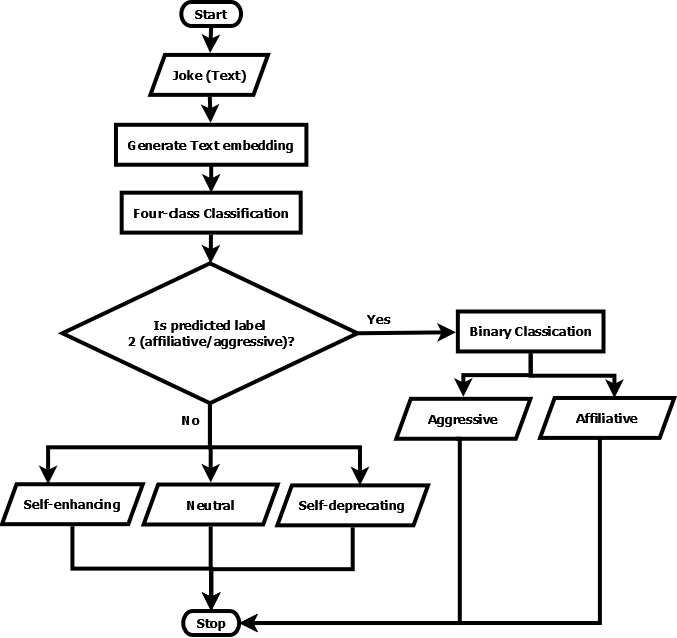}
  \caption{Flowchart illustrating the proposed Two-Model Approach for Humour Style Recognition}
  \label{fig:two_model_approach}
\end{figure}

This study employs two different approaches for humour style recognition: the single-model and the two-model approach. A total of 14 models were evaluated, including Naive Bayes, Random Forest, XGBoost (each with six different text embeddings), and DistilBERT. Figure \ref{fig:two_model_approach} illustrates the two-model approach, which first classifies humour instances into broader groups before refining to specific styles.

\subsection{Classifiers and Embedding Models}
\subsubsection{Classifiers}
The selection of classifiers was based on their suitability for the task at hand and efficiency in low-resource settings, avoiding resource-intensive large language models such as GPT4 and LLaMA prone to overfitting on small datasets due to their complex architectures \citep{Schur2024ComparativeModels,Diwakar2024DistilBERT-basedConditions,BerfuB2020AnalyzingClassification}:

\textbf{Naive Bayes (NB):} A probabilistic classifier based on the Bayes Theorem, assuming conditional independence of features given the target class \citep{Berrar2019BayesClassifier}.

\textbf{Random Forest (RF):} A bagging ensemble classifier using majority voting from multiple decision trees \citep{Jin2020ResearchDevelopment}.

\textbf{eXtreme Gradient Boosting (XGBoost):} A boosting ensemble classifier aggregating predictions of several weak learners, with regularisation to prevent overfitting \citep{Jiang2019AParameters}.

\textbf{DistilBERT:} A condensed BERT variant, offering faster performance and memory efficiency while maintaining competitive performance on NLP tasks \citep{Sanh2019DistilBERTLighter}.

\subsubsection{Sentence Embedding Models}
To capture distinct linguistic nuances and improve classification performance, we selected six embedding models from the top 20 on the Massive Text Embedding Benchmark (MTEB) leaderboard. These models were chosen for their robustness, efficiency, speed, and lightweight memory usage:

\begin{itemize}
    \item General Text Embeddings (GTE) and GTE Upgraded (ALI) \citep{Li2023TowardsLearning}
    \item BAAI General Embedding (BGE) \citep{Xiao2022RetroMAE:Auto-Encoder,Zhang2023RetrieveModels}
    \item Matryoshka Representation Learning and Binary Quantization (MRL) \citep{Lee2024Openemb2024mxbai}
    \item Universal AnglE Embedding (UAE) \citep{Li2023AnglE-optimizedEmbeddings}
    \item Multilingual E5 Text Embeddings (MUL) \citep{Wang2024MultilingualReport}
\end{itemize}

These embeddings were combined with RF and XGBoost classifiers for humour style recognition.

\subsection{Single-Model Approach}
In this approach, a single ML model is trained to classify the input text into one of the five classes: self-enhancing (\textbf{label 0}), self-deprecating (\textbf{label 1}), affiliative (\textbf{label 2}), aggressive (\textbf{label 3}), and neutral (\textbf{label 4}). This approach treats the task as a multi-class classification problem, where the model needs to distinguish between all five classes simultaneously.

\begin{figure*}[ht!]
    \centering
    \begin{minipage}{0.24\linewidth}
        \includegraphics[width=\linewidth]{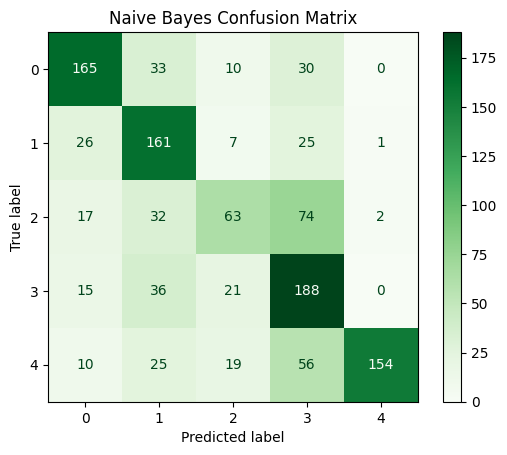}
        \subcaption{Naive Bayes }
        \label{fig:con_NB}
    \end{minipage}
    \hfill
    \begin{minipage}{0.24\linewidth}
        \includegraphics[width=\linewidth]{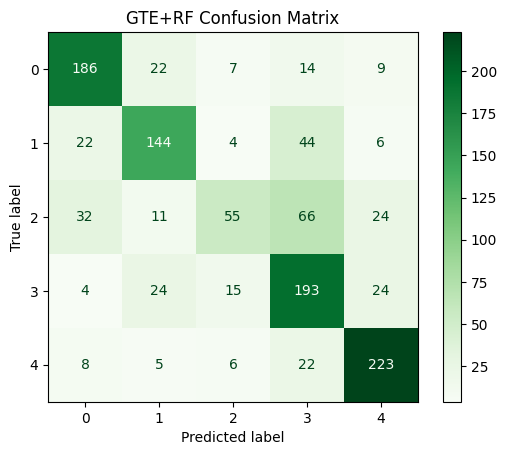}
        \subcaption{GTE+RF}
        \label{fig:con_GTE_RF}
    \end{minipage}
    \hfill
    \begin{minipage}{0.24\linewidth}
        \includegraphics[width=\linewidth]{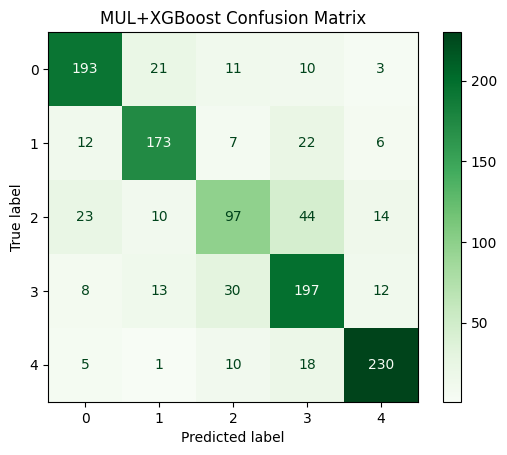}
        \subcaption{MUL+XGBoost}
        \label{fig:con_MUL_XGB}
    \end{minipage}
    \hfill
    \begin{minipage}{0.24\linewidth}
        \includegraphics[width=\linewidth]{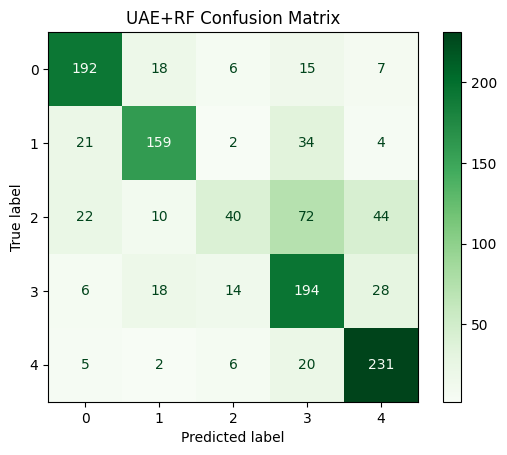}
        \subcaption{UAE+RF}
        \label{fig:con_UAE_RF}
    \end{minipage}
    \caption{5-Fold Cross Validation Confusion Matrix}
    \label{Fig:Confusion_Matrix}
\end{figure*}

To provide insight into the single-model performance, Figure \ref{Fig:Confusion_Matrix} presents the confusion matrices for the 5-fold cross-validation results of four models: Naive Bayes (NB) \ref{fig:con_NB}, GTE+RF \ref{fig:con_GTE_RF}, MUL+XGBoost \ref{fig:con_MUL_XGB}, and UAE+RF \ref{fig:con_UAE_RF}.

\subsection{Two-Model Approach}
To address limitations observed in the single-model approach, particularly in distinguishing affiliative humour, we developed a two-model approach. This method, inspired by previous studies \citep{Khan2022AClassification,VanLam2011Two-stagePages,Demidova2021TwostageAlgorithms}, improves classification performance by breaking down the problem into multiple steps.

The rationale behind this approach is to first separate the instances into broader groups and then focus on the more challenging task of distinguishing between affiliative and aggressive humour styles. This strategy is informed by an analysis of misclassified samples from the cross-validation and test set evaluation of the single-model approach, which revealed that affiliative humour was predominantly misclassified as aggressive humour. This pattern of misclassification is clearly illustrated in the cross-validation confusion matrices shown in Figure \ref{Fig:Confusion_Matrix}.  

The two-model approach involves two sequential steps:
\begin{enumerate}
    \item \textbf{Step 1: Four-Class Classification Model:} Train an ML model to distinguish between self-enhancing, self-deprecating, neutral, and a combined affiliative/aggressive class.
    \item \textbf{Step 2: Binary Classification Model:} Train a separate binary classification model to distinguish between affiliative and aggressive instances from the combined class in step 1.
\end{enumerate}

This approach allows for optimising overall performance by combining the best-performing models for each subtask.

\subsection{Experimental Setup}
The humour styles dataset was split 80/20 for training and testing, randomised using a fixed seed of 100 to ensure reproducibility. We used 5-fold cross-validation for all experiments to validate model performance and prevent overfitting.
For the NB classifier, we used a smoothing parameter of 1. The RF and XGBoost classifiers were implemented using their default hyperparameters. The DistilBERT model was fine-tuned for 5 epochs with a weight decay of 0.01, warmup steps of 500, and a training batch size of 8, using the default learning rate scheduler provided by the Hugging Face Transformers library.

\subsection{Evaluation Metrics}
Model performances were evaluated using standard metrics: accuracy, precision, recall, and f1-score. Accuracy measures overall performance, precision quantifies the ratio of true positives to predicted positives, recall assesses the model's ability to identify actual positives, and f1-score represents the harmonic mean of precision and recall. Furthermore, the Wilcoxon signed-rank test was used to compare the single-model and two-model approaches, determining the statistical significance of the performance differences between these approaches.

\section{Results and Discussions}
Experiments for the single-model and two-model approaches were conducted on Fourteen models: NB, RF + six embedding models, XGBoost + six embedding models and DistilBERT. 

\subsection{Baseline Model (Single-Model Approach)}
Tables \ref{Table:cross-validation} and \ref{Table:cross-validation-f1-score} show the mean accuracy and macro-mean f1-score of the 5-fold cross-validation for different models and embedding techniques, respectively. The results highlight the robustness and generalisability of our models across different data splits.

\begin{table*}[ht!]
    \centering
    \resizebox{1.0\textwidth}{!}{
        \begin{tabular}{l|c|cccccc|cccccc|c} \hline
        
            \textbf{Model} & \multicolumn{1}{|c|}{\textbf{NB (\%)}}   & \multicolumn{6}{c|}{\textbf{Random Forest (\%)}} & \multicolumn{6}{c|}{\textbf{XGBoost (\%)}} & \multicolumn{1}{c}{\textbf{DistilBERT (\%)}} \\ \hline
              
             &   & \textbf{BGE} & \textbf{GTE} &  \textbf{UAE} & \textbf{MRL} & \textbf{ALI} &  \textbf{MUL} & \textbf{BGE} & \textbf{GTE} &  \textbf{UAE} & \textbf{MRL} & \textbf{ALI} &  \textbf{MUL} & \\ \hline
            
            \textbf{Five-Class} & 62.5 & 69.2 & 68.5 & 69.7 & 67.0 & 70.4 &  71.9 & 69.7 & 71.2 &  72.1& 71.3 & 73.0 &  76.1 & 75.9 \\
            
            \textbf{Four-Class} & 66.0 & 75.4 & 74.3 &  74.3 & 72.8 & 76.5 & 79.1 &78.3  & 78.2 &  79.8 & 79.1 & 78.8 &  82.1 & 82.4  \\
            
            \textbf{Binary-Class} & 74.8 & 73.9 & 78.8 & 75.9 & 74.8 & 77.2 & 78.1 & 71.9 & 79.5 &  74.1 & 75.2 & 76.6 &  80.3 & 78.3 \\ \hline
            
        \end{tabular}
    }
    \caption{Mean Accuracy of 5-Fold Cross-Validation for the Various Classification Models}
    \label{Table:cross-validation}
\end{table*}

\begin{table*}[ht!]
    \centering
    \resizebox{1.0\textwidth}{!}{
        \begin{tabular}{l|c|cccccc|cccccc|c} \hline
        
            \textbf{Model} & \multicolumn{1}{|c|}{\textbf{NB (\%)}}   & \multicolumn{6}{c|}{\textbf{Random Forest (\%)}} & \multicolumn{6}{c|}{\textbf{XGBoost (\%)}} & \multicolumn{1}{c}{\textbf{DistilBERT (\%)}} \\ \hline
              
             &   & \textbf{BGE} & \textbf{GTE} &  \textbf{UAE} & \textbf{MRL} & \textbf{ALI} &  \textbf{MUL} & \textbf{BGE} & \textbf{GTE} &  \textbf{UAE} & \textbf{MRL} & \textbf{ALI} &  \textbf{MUL} & \\ \hline
            
            \textbf{Five-Class} & 61.4 & 65.2 & 65.9 & 65.9 & 63.5 & 67.9 &  69.0 & 67.7 & 70.1 &  71.0 & 70.1 & 71.6 & 74.9 & 74.6 \\
            
            \textbf{Four-Class} & 63.7 & 73.1 & 72.5 &  73.5 & 71.66 & 75.3 & 77.9 & 77.2 & 77.9 &  79.6 & 78.7 & 78.2 &  82.0 & 81.9  \\
            
            \textbf{Binary-Class} & 74.1 & 71.4 & 77.1 & 73.8 & 72.9 & 75.6 & 76.1 & 70.0 & 78.8 &  73.1 & 73.8 & 75.4&  79.5 & 77.7 \\ \hline
            
        \end{tabular}
    }
    \caption{Macro-mean F1-Score of 5-Fold Cross-Validation for the Various Classification Models}
    \label{Table:cross-validation-f1-score}
\end{table*}

Table \ref{Table:5_class_models} presents the overall performance for the five-class classification. MUL+RF, ALI+RF, and DistilBERT performed best with accuracies and f1-scores of 77.1\% and 76.6\%, 77.8\% and 77.3\%, and 75.4\% and 75.2\%, respectively.

\begin{table*}[ht!]
    \centering
    \resizebox{1.0\textwidth}{!}{
        \begin{tabular}{l|c|cccccc|cccccc|c} \hline
        
            & \multicolumn{1}{|c|}{\textbf{NB (\%)}}   & \multicolumn{6}{c|}{\textbf{Random Forest (\%)}} & \multicolumn{6}{c|}{\textbf{XGBoost (\%)}} & \multicolumn{1}{c}{\textbf{DistilBERT (\%)}} \\ \hline
              
             &   & \textbf{BGE} & \textbf{GTE} &  \textbf{UAE} & \textbf{MRL} & \textbf{ALI} &  \textbf{MUL} & \textbf{BGE} & \textbf{GTE} &  \textbf{UAE} & \textbf{MRL} & \textbf{ALI} &  \textbf{MUL} & \\ \hline
            
            \textbf{Precision} & 64.1 & 72.7 & 71.4 & 76.5 & 65.0 & 72.9 & 72.7 
            
            & 73.6 & 70.1 &  73.7 & 68.5 & \textbf{77.6} &  76.8 & 75.6 \\
            
            \textbf{Recall}   & 62.5 & 70.3 & 71.6 &  74.3 & 64.0 & 72.7 & 72.1 
            
            & 74.0 & 70.6 &  72.7 & 68.4 & \textbf{77.6} &  77.4 & 75.1 \\
            
            \textbf{F1-score}  &61.4 & 68.5 & 69.2 &  72.6 & 61.7 & 71.8 & 70.8 
            
            & 72.6 & 69.7 &  72.3 & 67.6 & \textbf{77.3} & 76.6 & 75.2 \\
            
            \textbf{Accuracy}  & 61.8 & 70.3 & 71.7 &  74.4 & 64.5 & 73.0 & 72.7 
            
            & 73.7 & 71.3 &  73.0 & 68.9 & \textbf{77.8} &  77.1 & 75.4 \\ \hline
            
        \end{tabular}
    }
    \caption{Performance of the Single-Model Approach}
    \label{Table:5_class_models}
\end{table*}

While the single-model approach achieved decent overall performance, Table \ref{Table:5_individual_class_f1score} reveals that all models struggle to identify affiliative humour accurately. Despite high overall accuracy, this approach fails to differentiate affiliative humour from other styles, particularly aggressive humour, as shown in Figure \ref{Fig:Confusion_Matrix}, highlighting a critical issue.

\begin{table*}[ht!]
    \centering
    \resizebox{1.0\textwidth}{!}{
        \begin{tabular}{l|c|cccccc|cccccc|c} \hline
        
              & \multicolumn{1}{|c|}{\textbf{NB (\%)}}   & \multicolumn{6}{c|}{\textbf{Random Forest (\%)}} & \multicolumn{6}{c|}{\textbf{XGBoost (\%)}} & \textbf{DistilBERT(\%)} \\ \hline
              
            \textbf{ Humour Styles} &   & \textbf{BGE} & \textbf{GTE} &  \textbf{UAE} & \textbf{MRL} & \textbf{ALI} &  \textbf{MUL} & \textbf{BGE} & \textbf{GTE} &  \textbf{UAE} & \textbf{MRL} & \textbf{ALI} &  \textbf{MUL} &\\ \hline
            
            \textbf{Self-enhancing} & 61.7& 80.3& 81.9 & 82.8 & 70.7 & 76.9 & 85.0 & 80.3 & 81.6 & 80.0 & 73.2 & 82.6 & \textbf{86.2}  &  79.4\\
            
            \textbf{Self-deprecating}  & 66.0 & 72.5 & 76.7 & \textbf{80.5} & 65.9 & 70.5 & 66.7 & 77.1 & 67.4 & 75.9 & 71.3 & 79.1 & 77.6  &  76.7 \\
            
            \textbf{Affiliative}   & \textcolor{red}{39.2} & \textcolor{red}{40.5} & \textcolor{red}{34.9} & \textcolor{red}{46.5}  & \textcolor{red}{33.7} & \textcolor{red}{54.5} & \textcolor{red}{47.3} & 
            \textcolor{red}{50.0} & \textcolor{red}{48.5} & \textcolor{red}{57.4} & \textcolor{red}{48.0} & \textcolor{red}{\textbf{64.9}} & \textcolor{red}{63.0}  & \textcolor{red}{60.2}\\
            
            \textbf{Aggressive} & 56.4 & 62.7 & 69.1 & 72.0 & 58.9 & 71.3 & 67.6  & 67.2 & 65.6 & 66.7 & 62.8 & \textbf{74.8} &  67.7 &  70.8 \\
            
            \textbf{Neutral}  & 83.6 & 86.3 & 83.4 & 81.3 & 79.5 & 85.7 & 87.1  
            & 88.2 & 85.3 & 81.6 & 82.6 & 85.1 & \textbf{88.7}  & \textbf{88.7} \\ \hline
            
        \end{tabular}
    }
    \caption{Macro-mean F1-score for each humour style for the Single-Model Approach}
    \label{Table:5_individual_class_f1score}
\end{table*}

This misclassification may stem from affiliative humour sometimes containing slightly aggressive components, as noted by \citet{Martin2003IndividualQuestionnaire}. For example:
\textit{\textbf{JOKE:} `To be happy with a man, you must understand him a lot and love him a little. To be happy with a woman, you must love her a lot and not try to understand her at all'.} (\textbf{LABEL: }\textit{\textcolor{blue}{True:`Affiliative'}, \textcolor{red}{Predicted: `Aggressive'}})

This joke attempts to playfully highlight gender differences, aiming for camaraderie. However, its misclassification as aggressive likely stems from the presence of gender stereotypes that could be misconstrued as demeaning. This example illustrates how subtle nuances in tone, context, and intent can lead to misclassifications between affiliative and aggressive humour. 

\subsection{Two-Model Approach}
To address the challenge of misclassifying affiliative humour as aggressive, we implemented a two-model approach, consisting of a four-class model and a binary-class model. The performance of these individual models is presented in Tables \ref{Table:4_and_2_individual_accuarcy} and \ref{Table:4_and_2_individual_f1-score}, which show their accuracy and macro-mean f1-score, respectively. Among the four-class models, MUL+XGBoost achieved the highest performance, with an accuracy of 85.3\% and a macro-mean f1-score of 85.1\%. In contrast, the binary-class model ALI+XGBoost outperformed the other models, with an accuracy and f1-score of 80.0\%.

The results of the two-model approach, which combines the four-class and binary models, are presented in Tables \ref{Table:two_model_approach} and \ref{Table:two_model_approach_f1score}. This approach yields improved overall performance compared to the single-model method, with the best results achieved by the combination of MUL+XGBoost and ALI+XGBoost, which attained a f1-score of 78.0\% and an accuracy of 77.8\%. Notably, in Tables \ref{Table:two_model_approach} and \ref{Table:two_model_approach_f1score}, MUL+XGBoost was consistently used as the four-class model in combination with various binary models (embeddings + RF or XGBoost), as it had previously demonstrated the best performance among the four-class models.

\begin{table*}[ht!]
    \centering
    \resizebox{1.0\textwidth}{!}{
        \begin{tabular}{l|c|cccccc|cccccc|c} \hline
        
              & \multicolumn{1}{|c|}{\textbf{NB (\%)}}   & \multicolumn{6}{c|}{\textbf{Random Forest (\%)}} & \multicolumn{6}{c|}{\textbf{XGBoost (\%)}} & \textbf{DistilBERT} (\%)\\ \hline
              
             \textbf{Models} &   & \textbf{BGE} & \textbf{GTE} &  \textbf{UAE} & \textbf{MRL} & \textbf{ALI} &  \textbf{MUL} & \textbf{BGE} & \textbf{GTE} &  \textbf{UAE} & \textbf{MRL} & \textbf{ALI} &  \textbf{MUL} & \\ \hline
            
            \textbf{Four-Class}     & 73.0 & 76.5 & 80.5 & 77.1 & 75.1 & 80.9& 83.6 & 80.5 & 80.9 & 81.2 & 76.8 & 82.6 & \textbf{85.3} & 82.6 \\ \hline
            
            \textbf{Binary-Class}   & 76.7 & 70.0 & 74.2 & 74.2 & 73.3 & 75.8 & 78.3 & 71.7 & 71.7 & 74.2 & 70.8 & \textbf{80.0} & 78.3 & 79.2 \\ \hline
            
        \end{tabular}
    }
    \caption{Performance Accuracy of Four-Class and Binary-Class Individual Models}
    \label{Table:4_and_2_individual_accuarcy}
\end{table*}

\begin{table*}[ht!]
    \centering
    \resizebox{1.0\textwidth}{!}{
        \begin{tabular}{l|c|cccccc|cccccc|c} \hline
        
              & \multicolumn{1}{|c|}{\textbf{NB (\%)}}   & \multicolumn{6}{c|}{\textbf{Random Forest (\%)}} & \multicolumn{6}{c|}{\textbf{XGBoost (\%)}} & \textbf{DistilBERT} (\%)\\ \hline
              
             \textbf{Models} &   & \textbf{BGE} & \textbf{GTE} &  \textbf{UAE} & \textbf{MRL} & \textbf{ALI} &  \textbf{MUL} & \textbf{BGE} & \textbf{GTE} &  \textbf{UAE} & \textbf{MRL} & \textbf{ALI} &  \textbf{MUL} & \\ \hline
            
            \textbf{Four-Class}     & 70.5 & 73.2 & 78.4 & 76.1 & 73.5 & 80.1 & 82.4 & 79.4 & 78.8 & 80.3 & 75.7 & 81.3 & \textbf{85.1} & 81.8 \\ \hline
            
            \textbf{Binary-Class}   & 76.3 & 69.5 & 73.9 & 73.8 & 73.1 & 75.8 & 78.3 & 71.6 & 71.4 & 73.9 & 70.7 & \textbf{80.0} & 78.3 & 79.2 \\ \hline
            
        \end{tabular}
    }
    \caption{Macro-mean F1-score of Four-Class and Binary-Class Individual Models}
    \label{Table:4_and_2_individual_f1-score}
\end{table*}

\begin{table*}[ht!]
    \centering
    \resizebox{1.0\textwidth}{!}{
        \begin{tabular}{l|c|cccccc|cccccc|c} \hline
        
              \textbf{Four-Class Model ->} & \multicolumn{1}{|c|}{\textbf{NB}}   & \multicolumn{6}{c|}{\textbf{MUL + XGBoost }} & \multicolumn{6}{c|}{\textbf{MUL + XGBoost}} & \textbf{DistilBERT} \\ \hline

             \multirow{2}{*}{\textbf{Binary-Class Model ->}} & \multirow{2}{*}{\textbf{NB (\%)}} & \multicolumn{6}{c|}{\textbf{Random Forest (\%)}} & \multicolumn{6}{c|}{\textbf{XGBoost (\%)}} &  \multirow{2}{*}{\textbf{DistilBERT (\%)}} \\ \cline{3-14}
              
             &   & \textbf{BGE} & \textbf{GTE} &  \textbf{UAE} & \textbf{MRL} & \textbf{ALI} &  \textbf{MUL} & \textbf{BGE} & \textbf{GTE} &  \textbf{UAE} & \textbf{MRL} & \textbf{ALI} &  \textbf{MUL} & \\ \hline

            \textbf{Precision} & 72.7 & 75.0 & 77.9 & 78.1 & 78.3 & 78.6 & 78.5 &  75.5 & 76.3 & 77.2 &  76.5 & \textbf{78.6} & 78.2 &  76.8  \\
            
            \textbf{Recall}   &  67.2 & 73.6 & 76.3 & 76.3 & 76.3 & 76.9 & 77.2 &  74.5 & 74.9 & 75.9 & 74.9  & \textbf{77.8} & 77.5 &  74.8    \\
            
            \textbf{F1-score} &  67.4  & 73.5  & 76.3 & 76.2 & 76.3 & 77.1 & 77.4 &  74.8 & 75.0 & 75.9 & 75.1 & \textbf{78.0} & 77.7 &  75.3 \\
            
            \textbf{Accuracy}  & 67.6 & 73.4 & 76.1 & 76.1 & 76.1 & 76.8 & 77.1 & 74.4  & 74.7 & 75.8 & 74.7 & \textbf{77.8} & 77.5 &  75.4  \\ \hline
             
        \end{tabular}
    }
    \caption{Performance of the Two-Model Approach}
    \label{Table:two_model_approach}
\end{table*}

\begin{table*}[ht!]
    \centering
    \resizebox{1.0\textwidth}{!}{
        \begin{tabular}{l|c|cccccc|cccccc|c} \hline
        
            \textbf{Four-Class Model ->} & \multicolumn{1}{|c|}{\textbf{NB}}   & \multicolumn{6}{c|}{\textbf{MUL + XGBoost }} & \multicolumn{6}{c|}{\textbf{MUL + XGBoost}} & \textbf{DistilBERT} \\ \hline

             \multirow{2}{*}{\textbf{Binary-Class Model ->}} & \multirow{2}{*}{\textbf{NB (\%)}} & \multicolumn{6}{c|}{\textbf{Random Forest (\%)}} & \multicolumn{6}{c|}{\textbf{XGBoost (\%)}} &  \multirow{2}{*}{\textbf{DistilBERT (\%)}} \\ \cline{3-14}
              
              &   & \textbf{BGE} & \textbf{GTE} &  \textbf{UAE} & \textbf{MRL} & \textbf{ALI} &  \textbf{MUL} & \textbf{BGE} & \textbf{GTE} &  \textbf{UAE} & \textbf{MRL} & \textbf{ALI} &  \textbf{MUL} & \\ \hline
            
            \textbf{Self-enhancing}   & 56.8 & 86.4 & 86.4 & 86.4 & 86.4 & 86.4 & 86.4 & 86.4  & 86.4 & 86.4 & 86.4 & 86.4 & 86.4 &  80.3 \\
            
            \textbf{Self-deprecating} & 66.7 & 80.9 & 80.9 & 80.9 & 80.9 & 80.9 & 80.9 & 80.9 & 80.9 & 80.9 & 80.9 & 80.9 & 80.9 &  75.6\\
            
            \textbf{Affiliative}      & 67.6 & 50.5 & 59.8 & 59.0 & 61.0 & 66.7 & 65.5 & 57.1  & 56.4 & 58.2 & 58.2 & \textbf{66.1} & 63.9 &  61.2 \\
            
            \textbf{Aggressive}       & 64.0 & 61.5 & 65.7 & 66.2 & 64.8 & 63.3 & 65.7 & 61.1  & 62.9 & 65.7 & 61.4 & 68.2 & 68.8 &  \textbf{71.2} \\
                      
            \textbf{Neutral}          & 81.7 & 88.4 & 88.4 & 88.4 & 88.4 & 88.4 & 88.4 & 88.4 & 88.4 & 88.4 & 88.4 & 88.4 & 88.4 & 88.1 \\ \hline
            
        \end{tabular}
    }
    \caption{Macro-mean F1-score for each humour style for the Two-Model Approach}
    \label{Table:two_model_approach_f1score}
\end{table*}

\begin{table*}[ht!]
\centering
    \resizebox{1.0\textwidth}{!}{
        \begin{tabular}{l|lllllllll} \hline
            & \textbf{Precision} & \textbf{Recall} & \textbf{F1-Score} & \textbf{Accuracy} & \textbf{Self-enhancing} & \textbf{Self-deprecating} & \textbf{Affiliative} & \textbf{Aggressive} & \textbf{Neutral} \\ \hline
            \textbf{Wilcoxon Statistics} & 0.0 & 3.0& 0.0 & 0.0  & 8.0 & 3.0  & 0.0 & 27.0 & 10.0 \\
            \textbf{P-value} & 0.000122  & 0.000610 & 0.000122 & 0.00220  & 0.0031  & 0.0006  & 0.0001 & 0.1189 & 0.0052 \\
            \textbf{Average (Single-Model)} & 72.23  & 71.66 & 70.52 & 71.83 & 78.76 & 73.14 & 49.19 & 66.69 & 84.79 \\
            \textbf{Average (Two-Model)} & 77.01& 75.29 & 75.43 & 75.25 & 83.85 & 79.51 & 60.80 & 65.04 & 87.90 \\
            \textbf{Model Difference} & 4.79  & 3.63 & 4.91 & 3.42 & 5.09 & 6.37 & 11.61 & -1.65 & 3.11 \\
            \textbf{\# of improved models out of 14} & 14 & 13 & 14 & 12 & 13 & 13 & 14 & 4 & 11 \\ \hline
    \end{tabular}
    }
    \caption{Wilcoxon Sign-Rank Test to Compare the Single-model and Two-model Approaches}
    \label{Table:wilcoxon_test}
\end{table*}

The Wilcoxon signed-rank test results (Table \ref{Table:wilcoxon_test}) statistically validate the improvements observed in the two-model approach. Significant improvements (p-value < 0.05) are evident for most metrics and humour styles, except aggressive humour (p-value = 0.1189). The two-model approach consistently outperforms the single-model approach, with average increases ranging from 3.42\% to 4.91\% across precision, recall, f1-score, and accuracy.

Notably, the two-model approach significantly improved affiliative humour classification, with an 11.61\% increase in f1-score. All 14 models showed improvement for affiliative humour under this approach, suggesting more robust and accurate classification, especially for previously challenging categories like affiliative humour.

The cross-validation results (Tables \ref{Table:cross-validation} and \ref{Table:cross-validation-f1-score}) further support the robustness of our findings. The five-class models' cross-validation accuracies and macro-mean f1-scores generally align with final test set accuracies and macro-mean f1-scores, indicating good generalisation. The four-class and binary-class models achieved even closer alignment, suggesting robust generalisation.

In summary, the two-model approach demonstrates superior performance in humour style recognition, particularly in identifying affiliative humour, with improved performance and generalisability across various metrics.

\section{Conclusion}
Automatic recognition of humour styles is a valuable yet challenging task with significant implications for digital humanities research, particularly in areas such as mental health, content moderation, and social media discourse. This study addresses the lack of established resources by introducing a new dataset of 1,463 instances across four humour styles and non-humour, while providing baseline evaluations of various models.

The dataset and research have significant implications in three key areas:
\begin{enumerate}
    \item \textbf{Mental Health}: Automatically identifying humour styles can enhance mental health research by enabling large-scale analysis of social media content. Different humour styles may correlate with various mental health indicators, potentially aiding in early detection of conditions such as depression or anxiety. For example, frequent use of self-deprecating humour might signal underlying mental health concerns.
    \item \textbf{Content Moderation}: The dataset can contribute to more refined content moderation systems on social media platforms. By distinguishing between different humour styles, moderators can better identify potentially harmful content disguised as humour, such as aggressive or self-defeating jokes, while allowing for benign forms of humour that enhance online interactions.
    \item \textbf{Social Media Discourse}: Automatic recognition of humour styles can provide valuable insights into social dynamics and communication patterns across various online communities. This can help researchers understand how different humour styles influence online discussions, shape public opinion, and contribute to the spread of information or misinformation.
\end{enumerate}

Our initial single-model approach struggled to accurately recognise affiliative humour, with f1-scores ranging from 39.2\% to 64.9\%. To address this, we developed a two-model approach consisting of a four-class model (merging affiliative and aggressive styles) followed by a binary model distinguishing between these styles. Extensive evaluation demonstrated the effectiveness of this approach in improving affiliative humour recognition, achieving f1-scores of 50.5\% to 66.1\%, while maintaining good performance for other styles. Furthermore, this approach offers flexibility in combining the best models for each sub-task, optimising overall performance.

By introducing this dataset and baseline evaluations, we aim to catalyse further research and development in these critical areas of digital humanities, ultimately enhancing our understanding of humour and its multifaceted impact on human communication.

\section{Dataset Availability}
The dataset and models implemented in this study are available to the community via the link in the footnote \footnote{Humour Styles Dataset: \url{https://github.com/MaryKenneth/Two_Model_Humour_Style}}. Additionally, thirty instances from the dataset are included in Appendix \ref{appendix: Sample Jokes Dataset}.

\section{Limitations and Future Works}
This study has several limitations. The dataset, consisting of 1,463 instances, is relatively small, which may limit the model's generalisation capabilities. Additionally, the inherent subjectivity of humour, along with the observed inter-rater agreement and annotation disagreements, underscores the challenges in consistently labelling humorous content. The focus on English-centric jokes may also introduce biases and language-specific nuances.

Future research could focus on collecting larger and more diverse datasets from various languages and sources to improve the robustness of the model. Leveraging transfer learning methods, such as intermediate fine-tuning on pre-trained language models, could enhance performance, especially when data is limited. Exploring multimodal approaches that incorporate visual, auditory, and contextual cues, as well as personalised models that adapt to individual preferences, could provide deeper insights into humour styles. Furthermore, investigating generative models for producing humorous content in specific styles presents a promising direction for further exploration.

Despite these limitations, this study lays the groundwork for humour style recognition, paving the way for extensive future research on computational humour analysis and its applications in digital humanities.

\section{Acknowledgement}
This research was supported by the Petroleum Technology Development Fund (PTDF) of Nigeria.

\bibliography{references}

\appendix

\section{Humour Style Websites}
\label{appendix:humour_websites}

The website sources for the different humour styles, along with their corresponding links, are listed in Table \ref{Table:dataset-website-links}.

\begin{table}[H]
  \centering
  \resizebox{0.4\textwidth}{!}{%
    \begin{tabular}{ll}
      \hline
      \textbf{Humour Styles} & \textbf{Website} \\ \hline
      \multirow{4}{*}{Aggressive} & \href{https://parade.com/1295709/marynliles/dark-humor-jokes/}{Parade} \\
      & \href{https://www.laughfactory.com/jokes/insult-jokes}{Laugh factory} \\
      & \href{https://www.rd.com/list/funny-insults/}{Reader's digest} \\
      & \href{https://pun.me/funny/insults/}{Pun Me} \\ \hline
      
      \multirow{5}{*}{Affiliative} & \href{https://www.rd.com/article/funny-friendship-quotes/}{Reader's digest}  \\
      & \href{https://www.independent.co.uk/news/uk/home-news/international-day-of-happiness-best-jokes-ever-possibly-9205244.html}{Independent} \\
      & \href{https://happynumbers.com/blog/30-funniest-jokes-for-math-teachers/}{Happy Numbers } \\
      & \href{https://www.laughfactory.com/jokes/family-jokes}{Laugh factory}  \\
      & \href{https://teambuilding.com/blog/icebreaker-jokes}{Team building} \\ \hline
      
      \multirow{5}{*}{Self-Deprecating} & \href{https://www.tastefullyoffensive.com/2021/03/dark-funny-self-deprecating-memes/}{Tastefully Offensive} \\ 
      
        & \href{https://www.boredpanda.com/self-deprecating-jokes/}{Bored Pandas} \\ 
        
        & \href{https://www.cracked.com/article_36985_15-of-the-best-self-deprecating-jokes-from-top-comedians.html}{Cracked}  \\
        & \href{https://www.reddit.com/r/Standup/comments/34kwuy/what_is_your_go_to_self_deprecation_joke/}{Reddit}  \\  
        & \href{https://www.buzzfeed.com/annakopsky/by-hating-yourself-most}{Buzz Feed} \\ 
        \hline

        \multirow{5}{*}{Self-Enhancing} & \href{https://putthekettleon.ca/funny-self-care-quotes/}{Put the Kettle On} \\ 
        & \href{https://www.silkandsonder.com/blogs/news/44-funny-self-love-quote-thatll-make-you-laugh}{Silk and Sonder}  \\
        & \href{https://carleyschweet.com/funny-self-care-quotes/}{Carley Schweet} \\  
        & \href{https://www.joyfulthroughitall.com/funny-self-love-quotes/}{Joyful through it all} \\ 
        & \href{https://lauraconteuse.com/funny-self-love-quotes/}{Laura Conteuse} \\ 
        \hline
    \end{tabular}
  }
  \caption{List of websites from which jokes were taken.}
  \label{Table:dataset-website-links}
\end{table}

\section{Mapping Jokes to Humour Style Labels}
\label{appendix:mapping_labels}

Although certain humour websites from which the jokes were extracted do not explicitly categorise the humour as "aggressive," "affiliative," or "self-enhancing," there are reasonable justifications for associating the humour found on those sites with the respective humour styles, based on the content and intended audience. This section outlines the keywords and rationale for mapping jokes to the original labels, definitions, or tags provided for jokes on the websites.

\subsection{Aggressive Humour}
Aggressive humour is characterised by jokes, insults, or humorous remarks that are intended to disparage, belittle, or target particular individuals or groups. This type of humour often involves sarcasm, mockery, and put-downs, and it can be perceived as offensive or hostile by the targeted parties.

\subsubsection{Equivalence classes (Website Keywords)}
\begin{itemize}
    \item \textbf{Dark (inappropriate) Jokes: } Dark (inappropriate) jokes are identified as being cruel, morbid, or offensive to some, which aligns with the characteristics of aggressive humour \citep{Tang2022TheHumor}.
    \item \textbf{Insult: } Insult is an offensive remark or action intended to mock or belittle the target (Cambrige Dictionary (\url{https://rb.gy/l0b2sz})). Insult is a key characteristic of aggressive humour \citep{Martin2003IndividualQuestionnaire}.
\end{itemize}

\subsection{Affiliative Humour}
Affiliative humour is characterised by jokes, witty remarks, or humorous anecdotes that are intended to amuse others, facilitate social interactions, and strengthen relationships. This type of humour is non-hostile, benign, and often used to create a positive, inclusive atmosphere.
\subsubsection{Equivalence classes (Website Keywords)}
\begin{itemize}
    \item \textbf{Icebreakers Jokes for Work Meetings: } This jokes are typically used to create a relaxed and friendly environment in professional or group settings.They are meant to facilitate social interactions and put people at ease, which aligns with the goals of affiliative humour \citep{Cooper2013JustStudents}.
    \item \textbf{International Day of Happiness: } Jokes shared on occasions like the International Day of Happiness are typically intended to spread positivity, joy, and laughter among people. Such jokes are designed to bring people together and create a shared experience of amusement, which aligns with the goals of affiliative humour
    \item \textbf{Friendship: } Jokes meant to be shared among friends are often used to strengthen bonds, create shared laughter experiences, and reinforce the positive aspects of friendship. This type of humour is non-threatening and aimed at building connections, which is a characteristic of affiliative humour.
    \item \textbf{Family jokes: } Jokes shared within families are often intended to create a sense of bonding, shared laughter, and enjoyment. Family jokes are generally non-offensive and serve to strengthen familial relationships, which is a characteristic of affiliative humour \citep{Hedin2012JokesSweden,Gyasi2023HumorPerspective}.
    \item \textbf{Classroom: }Humour shared between teachers and students, or within educational settings, is often used to create a positive and engaging learning environment. These jokes are likely meant to connect with students and foster a sense of camaraderie, which is in line with affiliative humour \citep{Deiter2000TheClassroom,Jeder2015ImplicationsClassroom}
\end{itemize}
      
\subsection{Self-deprecating Humour}
Self-deprecating humour is a type of humour in which individuals make fun of their own flaws, weaknesses, or mistakes. It involves mocking or belittling oneself in a humorous way. 
\subsubsection{Equivalence classes (Website Keywords)}
\begin{itemize}
    \item \textbf{Self-deprecating: } The title on the websites directly states that the quotes and captions are "self-deprecating," implying that they involve humour directed at oneself in a self-mocking or self-effacing manner. Given the explicit use of the term "self-deprecating" in the titles of the websites, the jokes found on these sites are labelled as self-deprecating jokes.
\end{itemize}

\subsection{Self-enhancing Humour}
Self-enhancing humour is characterised by jokes, witty remarks, or humorous anecdotes that are intended to promote a positive self-image, boost self-confidence, and enhance one's sense of self-worth. This type of humour often involves self-affirmation, playful boasting, or exaggerating one's positive qualities in a light-hearted and non-hostile manner.
\subsubsection{Equivalence classes (Website Keywords)}
\begin{itemize}
    \item \textbf{Self-love: } Self-love, involves deliberately prioritising oneself, supporting your needs and desires, and respecting your limitations. It entails refraining from self-criticism, regret, shame, or guilt, and confronting uncomfortable emotions rather than avoiding them \citep{ClevelandClinic2024Self-Love:Yourself}. Self-love is closely tied to self-enhancement, as it involves promoting a positive self-image and boosting self-confidence. Humorous texts that encourage self-love can be seen as self-enhancing, as they aim to make one feel better about themselves and promote self-acceptance.
    \item \textbf{Self-care: } Self-care is the intentional practice of dedicating time to activities that promote overall well-being, encompassing both physical and mental health benefits. By effectively managing stress, reducing the risk of illness, and increasing energy levels, self-care fosters a healthier lifestyle \citep{NationalInstituteofMentalHealth2024CaringHealth}. A key aspect of self-care is cultivating a positive self-image and nurturing one's own well-being. In this context, humorous texts that aim to promote self-affirmation and boost self-confidence can be seen as self-enhancing, as they seek to enhance an individual's self-worth and overall sense of well-being.
\end{itemize}

\section{Word Clouds of Humour Styles Phrases }
\label{appendix:word_cloud}
\begin{figure}[H]
    \centering
        \includegraphics[width=\linewidth]{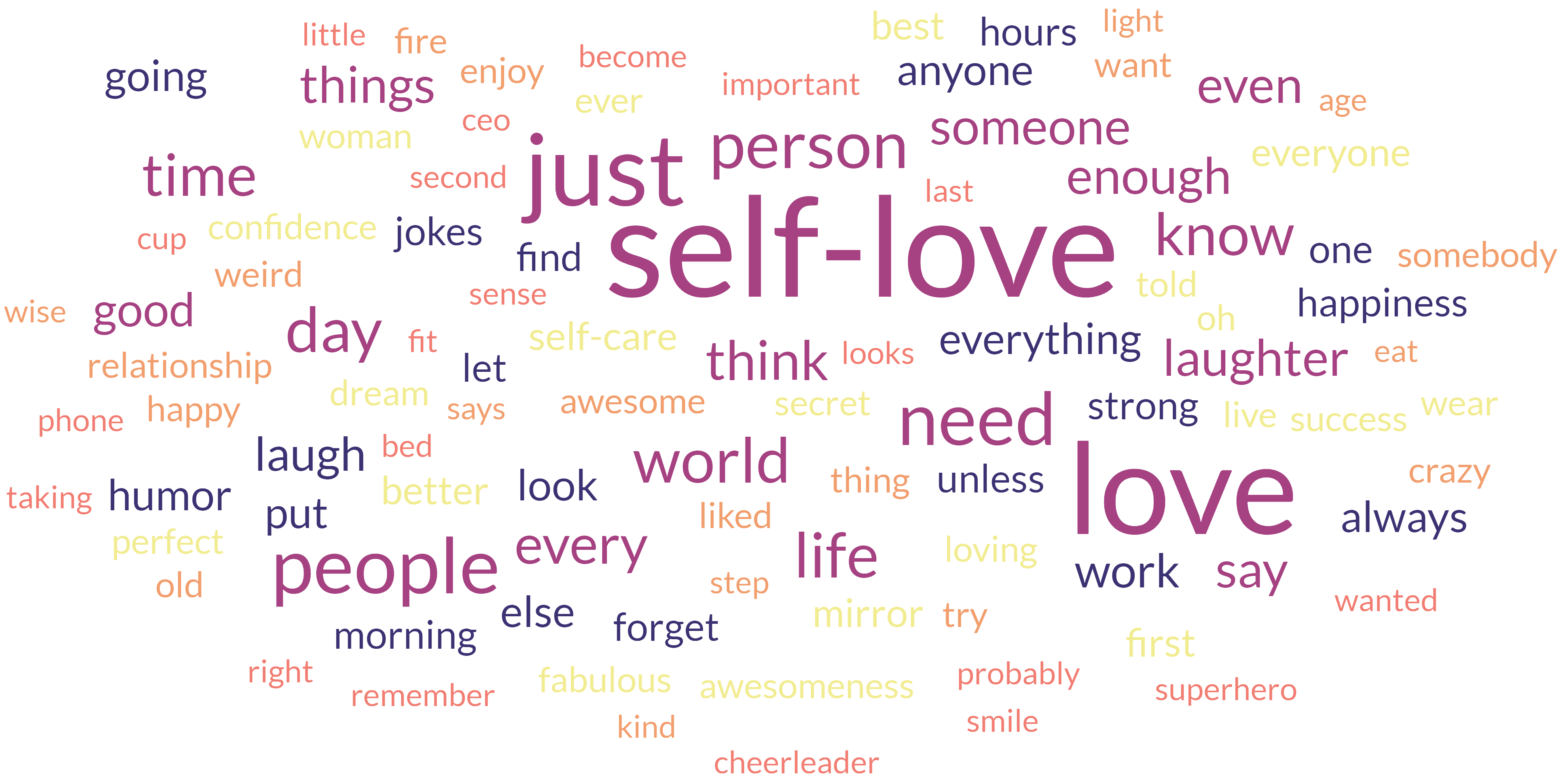}
        \caption{Most Frequent Self-Enhancing Phrases}
        \label{fig:self_enhancing_words}
\end{figure}

\begin{figure}[H]
    \centering
    \includegraphics[width=\linewidth]{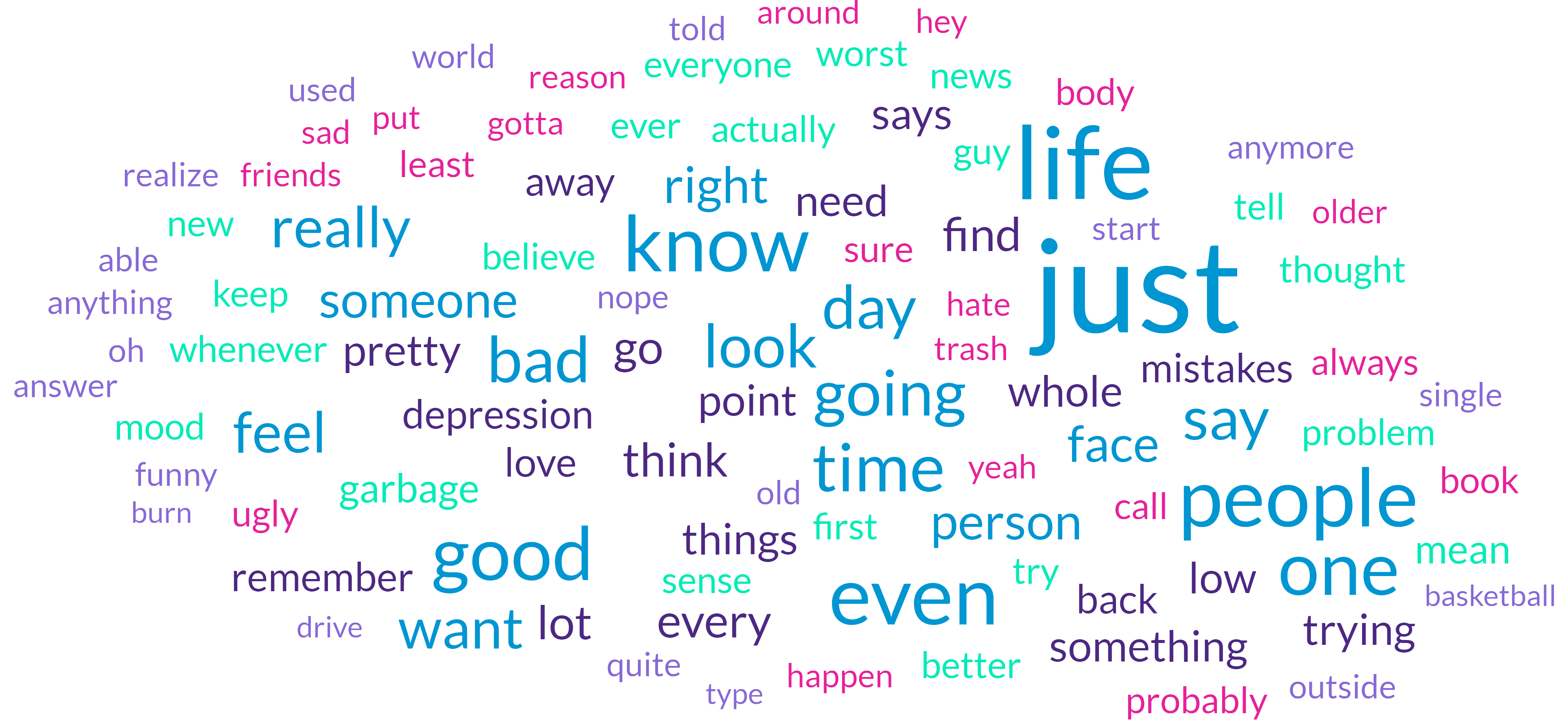}
    \caption{Most Frequent Self-Deprecating Phrases}
    \label{fig:self_deprecating_words}
\end{figure}

\begin{figure}[H]
    \centering
    \includegraphics[width=\linewidth]{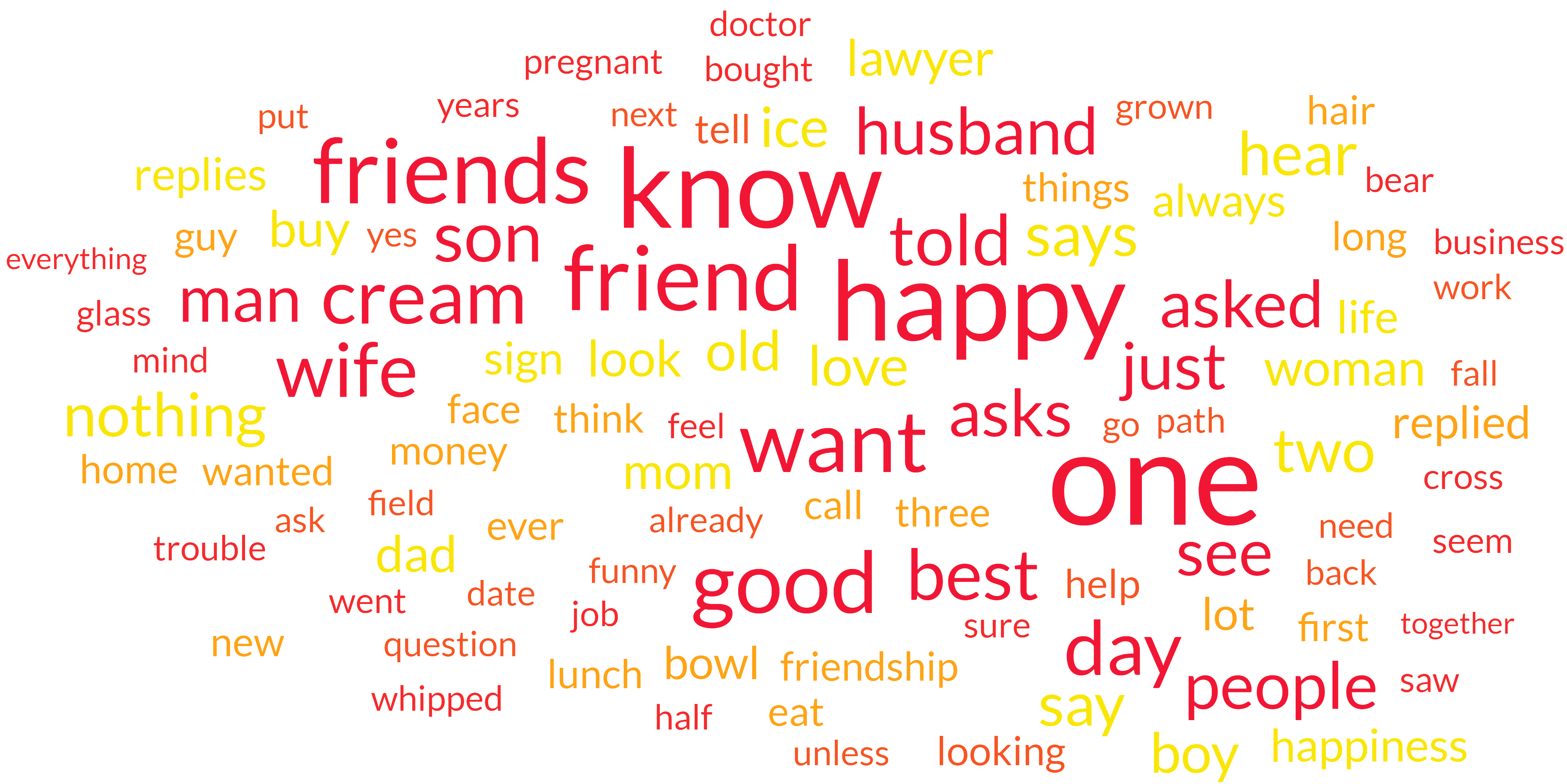}
    \caption{Most Frequent Affiliative Humour Phrases}
    \label{fig:affiliative_words}
\end{figure}

\begin{figure}[H]
    \centering
    \includegraphics[width=\linewidth]{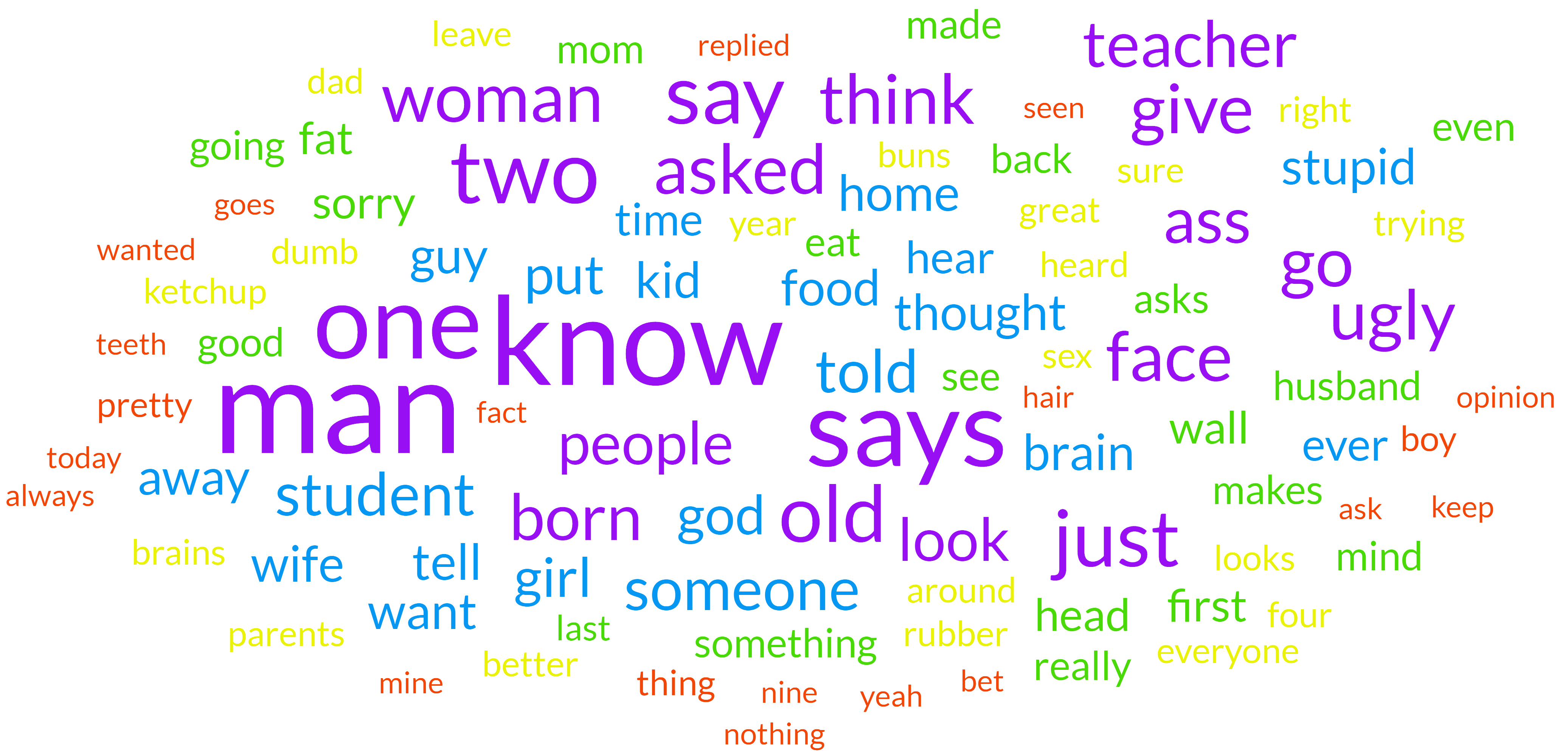}
    \caption{Most Frequent Aggressive Humour Phrases}
    \label{fig:aggressive_words}
\end{figure}

Figure \ref{fig:examples_styles} provides a selection of examples from the dataset for each humour style. Figures \ref{fig:self_enhancing_words}, \ref{fig:self_deprecating_words}, \ref{fig:affiliative_words}, and \ref{fig:aggressive_words} represent word clouds of the most common words associated with each of the humour styles in the created dataset. Figures \ref{fig:self_enhancing_words} and \ref{fig:affiliative_words} reveal a prevalence of positive phrases, including self-love, laughter, good, love, friends, and happy, in self-enhancing and affiliative humour. In contrast, Figures \ref{fig:self_deprecating_words} and \ref{fig:aggressive_words} highlight the presence of negative phrases, including ugly, fat, stupid, bad, depression, and mistakes, in self-deprecating and aggressive humour styles. 


\section{Annotation Disagreement}
\label{appendix:annotation_disagreement}
Table \ref{tab:annotation_disagreement} presents examples of jokes where the three human annotators (A1, A2, and A3) for each of the jokes disagreed on the annotation labels, as discussed in Section 3.3. The labels are interpreted as follows: self-enhancing (0), self-deprecating (1), affiliative (2), aggressive (3), and neutral (4). To gain insight into the annotation process, each rater, along with the LLM models, was asked to provide a rationale for their label assignments. Below, we summarise the reasons behind the label assignments for four of these jokes.

\hfill \break
\textbf{JOKE:} \textit{Insanity is hereditary, - You get it from your children.}

The annotators had varying interpretations of this joke, which are summarised below:

\begin{itemize}
    \item \textbf{Affiliative -2 (Copilot, ChatGPT, and A3):} This joke is light-hearted and relatable, playing on the common experiences of parenting. It is inclusive and bonding, fostering a sense of shared understanding.
    \item \textbf{Self-enhancing -0 (A1 and Claude):} The joke-teller uses a playful and light-hearted tone to poke fun at themselves, without being overly self-critical. The joke does not come across as aggressive or hostile towards anyone.
    \item \textbf{Self-deprecating -1 (HuggingChat):} Although not aggressive or mocking, the joke can be seen as self-deprecating. It humorously comments on the challenges of parenting, implying that the joke-teller is not immune to the stresses of parenthood.
    \item \textbf{Aggressive -3 (A2):} In contrast, one annotator interpreted the joke as aggressive, believing that it mocks and belittles parents.
\end{itemize}

\hfill \break
\textbf{JOKE:} \textit{Don’t worry if you’re a kleptomaniac, you can always take something for it.}
\begin{itemize}
    \item \textbf{Affiliative -2 (Claude, Copilot, ChatGPT, and A3):} This joke uses a lighthearted and playful tone to make a humorous comment about kleptomania, potentially creating a sense of shared understanding and camaraderie. Its intention is to be humorous rather than offensive.
    \item \textbf{Self-enhancing -0 (A1):} The joke-teller attempts to reframe their mental health disorder in a positive light, presenting it in a humorous and optimistic way.
    \item \textbf{Self-deprecating -1 (HuggingChat):} The joke can be seen as self-deprecating, as it humorously acknowledges the potentially embarrassing or shameful nature of kleptomania.
    \item \textbf{Aggressive -3 (A2):} In contrast, one annotator interpreted the joke as aggressive, believing that it belittles and mocks individuals with kleptomania, a mental health disorder.
\end{itemize}

\hfill \break
\textbf{JOKE:} \textit{Always follow your dreams. Except for that one where you're naked at work.}
\begin{itemize}
    \item \textbf{Affiliative -2 (Copilot, ChatGPT, and A1):} This joke takes a common piece of advice and adds a humorous twist that many people can relate to. It aims to create a sense of shared amusement over a common anxiety.
    \item \textbf{Self-enhancing -0 (A2):} The speaker presents themselves as someone who can laugh at their own imperfections and embarrassing moments, showcasing a positive coping mechanism. 
    \item \textbf{Self-deprecating -1 (HuggingChat and Claude):} The joke uses self-deprecation to poke fun at the embarrassing nature of certain dreams, with the joke-teller willingly making themselves the target of the humour.
    \item \textbf{Neutral -4 (A3):} This joke is interpreted as a neutral observational joke, lacking strong emotional undertones and instead focusing on a humorous observation.
\end{itemize}

\hfill \break
\textbf{JOKE:} \textit{Never get stuck behind the Devil in a Post Office queue! The Devil can take many forms.}
\begin{itemize}
    \item \textbf{Neutral -4 (HuggingChat, Copilot, ChatGPT, and A3):} This joke is a lighthearted commentary on the frustrations of waiting in line, without any specific target or malicious intent.
    \item \textbf{Affiliative -2 (Claude and A1):} The joke creates a sense of shared understanding and camaraderie around the common experience of waiting in line, which most people can relate to.  
    \item \textbf{Aggressive (A2):} In contrast, one annotator interpreted the joke as aggressive, as the Devil's representation of negative traits or behaviours could be seen as a critique of people in general.
\end{itemize}

\begin{table*}[ht!]
    \centering
    \resizebox{0.9\textwidth}{!}{
        \begin{tabular}{p{0.55\linewidth}|p{0.04\linewidth}|p{0.04\linewidth}|p{0.04\linewidth}|p{0.08\linewidth}|p{0.08\linewidth}|p{0.08\linewidth}|p{0.06\linewidth}} \hline
             \textbf{Jokes} & \textbf{A1}  & \textbf{A2} & \textbf{A3} &  \textbf{Hugging Chat} & \textbf{Claude} & \textbf{Copilot} &  \textbf{Chat-GPT} \\ \hline
            
            Insanity is hereditary, - You get it from your children. & 0 & 3 & 2 & 1 & 0 & 2 & 2 \\ \hline
            Gravity doesn't exist: the earth sucks. & 0 & 4 & 3 & 4 & 0 & 4 & 4 \\ \hline
            Did you hear about the Scottish Kamikaze pilot? He crashed his plane in his brother's junkyard & 3 & 4 & 2 & 3 & 3 & 3 & 3 \\ \hline
            Biology grows on you & 4 & 3 & 2 & 2 & 4 & 4 & 4 \\ \hline
            Don't worry if you're a kleptomaniac, you can always take something for it & 0 & 3 & 2 & 1 & 2 & 2 & 2 \\ \hline
            To steal from one is plagiarism. To steal from many is research & 2 & 3 &4  & 1 & 0 & 2 & 2 \\ \hline
            If all else fails, lower your standards & 2 & 1 & 4 & 1 & 1 & 1 & 1 \\ \hline
            There are only 3 things that tell the truth: 1 - Young Children 2 - Drunks 3 - Leggings & 1 & 3 & 4 & 2 & 4 & 2 & 2 \\ \hline
            Never get stuck behind the Devil in a Post Office queue! The Devil can take many forms. & 2 & 3 & 4 & 4 & 2 & 4 & 4 \\ \hline
            Always follow your dreams. Except for that one where you're naked at work. & 2 & 0 & 4 & 1 & 1 & 2 & 2 \\ \hline
            
        \end{tabular}
    }
    \caption{Annotation Disagreement}
    \label{tab:annotation_disagreement}
\end{table*}

\section{Sample Jokes Dataset}
\label{appendix: Sample Jokes Dataset}

In this section, we showcase a random selection of thirty samples from our jokes dataset (see Table \ref{tab:dataset_samples}). Each sample consists of the joke content paired with its corresponding label, providing a glimpse into the dataset's composition and structure. For reference, the labels are interpreted as follows:
\begin{itemize}
    \item Self-enhancing: 0
    \item Self-deprecating: 1
    \item Affiliative: 2 
    \item Aggressive: 3
    \item Neutral: 4
\end{itemize}

\begin{table*}[ht!]
    \centering
    \resizebox{0.9\textwidth}{!}{
    \begin{tabular}{p{0.90\linewidth}|p{0.07\linewidth}}
    \hline
    \textbf{Jokes} & \textbf{Labels} \\
    \hline
    
    Is that your nose or are you eating a banana? & 3 \\ \hline
    Q: Why did the witches' team lose the baseball game? A: Their bats flew away. & 2 \\ \hline
    Act your age, not your shoe size. & 3 \\ \hline
    I may be trash, but I burn with a bright flame & 1 \\ \hline
    Yeah, I know. I hate me too. & 1 \\ \hline
    “The secret of staying young is to live honestly, eat slowly, and lie about your age.” & 0 \\ \hline
    “I got it all together. But I forgot where I put it.” & 0 \\ \hline
    A man on a date wonders if he'll get lucky. A woman already knows. & 2 \\ \hline
    Here's how unfair the tax system is in each state &	4 \\ \hline
    Is a death sentence really a death sentence? &	4 \\ \hline
    Trump's new military plan will cost 150 billion dollars -- at the very least &	4 \\ \hline
    He is so short, his hair smells like feet. & 3 \\ \hline
    You should be in commercials for birth control.	& 3 \\ \hline
    “The road to success is dotted with many tempting parking spaces.” &	0 \\ \hline
    If I had a face like yours, I'd sue my parents! &	3 \\ \hline
    “I’m not perfect, but I’m perfectly me.” &	0 \\ \hline
    Why don't scientists trust atoms? Because they make up everything! & 2 \\ \hline
    Don’t mind me. I’m just having an existential crisis. Move along, folks. &	1 \\ \hline
    I can't talk to you right now, tell me, where will you be in 10 years? &	3 \\ \hline
    A wise woman once said, “fuck this shit” and lived happily ever after. &	0 \\ \hline
    He is depriving a village somewhere of an idiot. &	3 \\ \hline
    Dad: "Can I see your report card, son?" Son: "I don't have it." Dad: "Why?" Son: "I gave it to my friend. He wanted to scare his parents." &	2 \\ \hline
    “The elevator to success is out of order. You’ll have to use the stairs, one step at a time.” & 0 \\ \hline
    “I’m not the kind of guy who has a huge weight problem, but I am the kind of guy who could really put the brakes on an orgy. Everyone would be like, ‘Was he invited? Why is he eating a cake?’ I’ve never been in an orgy, but I feel like it’d be like what happens when I try to play pickup basketball: No one passes me the ball, and everyone asks me to keep my shirt on.” &	1 \\ \hline
    “I’m a self-love junkie. Can’t get enough of this good stuff!” &	0 \\ \hline
    “If I could rearrange the alphabet, I’d put ‘U’ and ‘I’ together.” &	2 \\ \hline
    “Let your light shine bright so the other weirdos can’t find you”	& 0 \\ \hline
    Did you hear about the magic tractor? It turned into a field.	& 2 \\ \hline
    I don’t have a nervous system. I am a nervous system!	& 1 \\ \hline
    How to build muscle: proven strength lessons from milo of croton &	4 \\ \hline

    \end{tabular}
    }
    \caption{Samples from the Humour Styles Dataset}
    \label{tab:dataset_samples}
\end{table*}

\end{document}